\documentclass{article}
\usepackage{spconf,amsmath,graphicx}
\usepackage{algorithm}      
\usepackage{algorithmic}     
\usepackage{pdfpages}      
\usepackage{graphicx}
\usepackage{amssymb}
\usepackage{dsfont}
\usepackage{array}
\usepackage{threeparttable}
\usepackage{makecell}
\usepackage{float}
\usepackage{multirow}

\title{Online Filter Clustering and Pruning for Efficient Convnets}
\name{Zhengguang Zhou$^{1}$, Wengang Zhou$^{1}$, Richang Hong$^{2}$, Houqiang Li$^{1}$
\thanks{This work was supported in part by 973 Program under Contract 2015CB351803 and Natural Science Foundation of China (NSFC) under Contract 61632019 and 61390514 and the Fundamental Research Funds for the Central Universities.}}
\address{$^{1}$CAS Key Laboratory of Technology in Geo-spatial Information Processing and Application System, \\
EEIS Department, University of Science and Technology of China\\
$^{2}$HeFei University of Technology\\
zhgzh164@mail.ustc.edu.cn,  zhwg@ustc.edu.cn,  hongrc.hfut@gmail.com, lihq@ustc.edu.cn}

\begin{document}

\maketitle

%
\begin{abstract}
 Pruning filters is an effective method for accelerating deep neural networks (DNNs), but most existing approaches prune filters on a pre-trained network directly which limits in acceleration. Although each filter has its own  effect in DNNs, but if two filters are same with each other, we could prune one safely. In this paper, we add an extra cluster loss term in the loss function which can force filters in each cluster to be similar online. After training, we keep one filter in each cluster and prune others and fine-tune the pruned network to compensate the loss. Particularly, the clusters in every layer can be defined firstly which is effective for pruning DNNs within residual blocks. Extensive experiments on CIFAR10 and CIFAR100 benchmarks demonstrate the competitive performance of our proposed filter pruning method.
\end{abstract}

\section{Introduction}
Deep neural networks (DNNs) have achieved state-of-art performance in many computer vision tasks \cite{liu20163d}\cite{khalaf2016convolutional} and grown deeper and deeper. However, these high capacity networks suffer high complexity in both storage and computation especially when used in resource-limited platforms, such as mobile phones and embedded devices \cite{sun2016scalable}. Thus, many researchers have a significant interest in network compression methods for reducing the storage and computation costs.

With the observation that DNNs have a significant parameter redundancy, pruning methods have been widely studied for reducing the number of parameters in networks. In the earlier years, researchers prune weights in a network, but it has a limitation in accelerating network inference \cite{lecun1990optimal}\cite{han2015learning}. Afterwards, more and more works focus on pruning filters or channels which can result in a thinner model and significant accelerations \cite{zhou1226}\cite{he2017channel}\cite{hu2016network}\cite{li2016pruning}\cite{liu2017learning}\cite{luo2017thinet}\cite{molchanov2016pruning}\cite{srinivas2015data}. The main step of the above pruning methods is to measure the importance score of each filter. After that, they prune the least important filters followed by a fine-tuning process to recover the performance of the network.
Some methods are based on the magnitude of the filter \cite{li2016pruning} and some approaches exploit the information of feature maps to estimate the importance of the corresponding
 filter \cite{he2017channel}\cite{luo2017thinet}. Moreover, if two weights are similar, one of them is redundant and can be pruned \cite{srinivas2015data}, but it only considers the fully-connected layers and limits in accelerating convolutional neural networks.

\begin{figure}[!t]
\centering
\includegraphics[scale = 0.7]{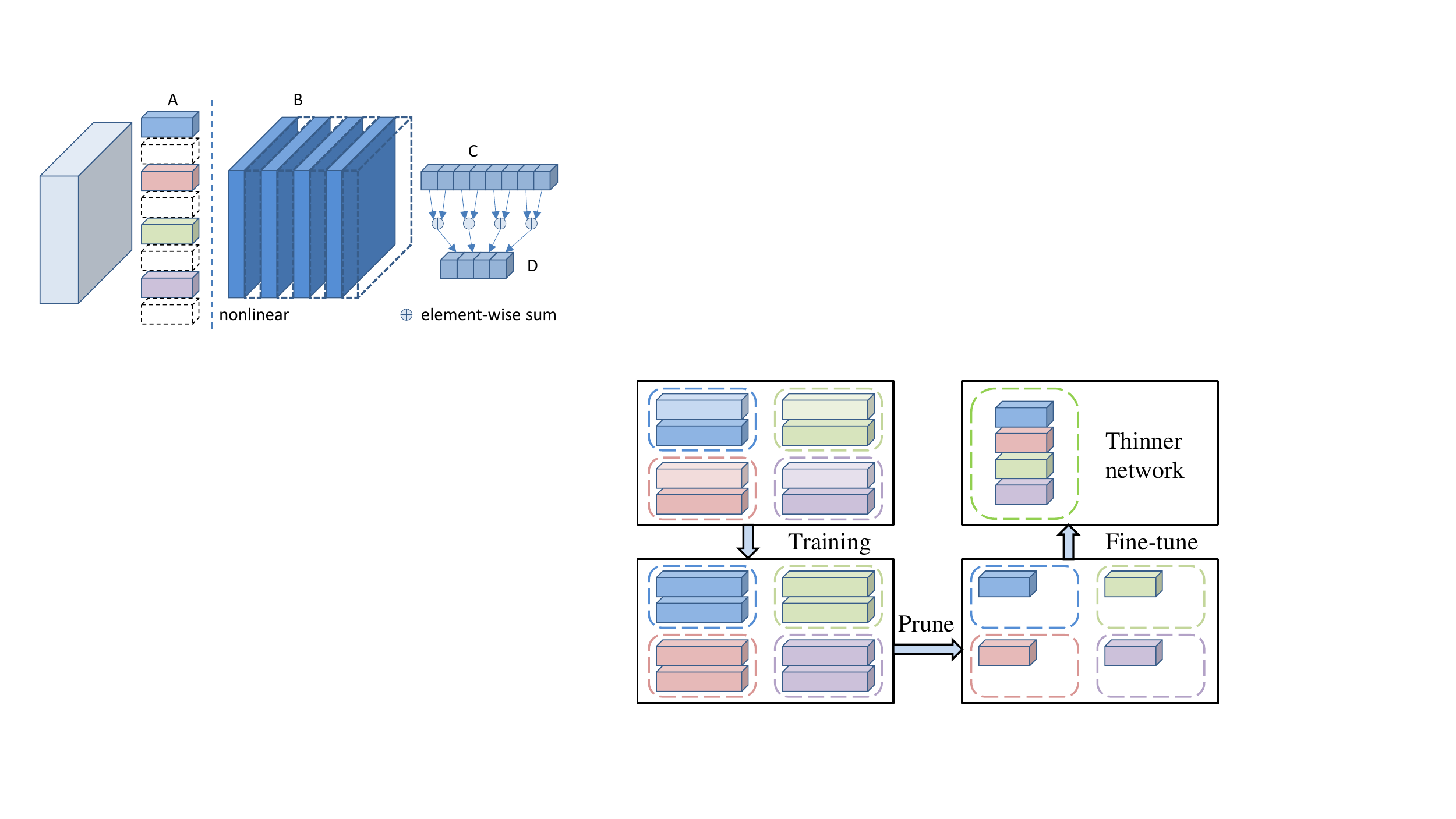}
\caption{The flow chart of our proposed method (\emph{e.g.,} a convolutional layer including 8 filters and 4 clusters).}
\label{filter_cluster}
\end{figure}

In this paper, we propose a new filter level pruning method.  Our method is based on the fact that if two filters are similar, one of them is redundant and can be effectively removed \cite{srinivas2015data}.   But two filter sets are always dissimilar in a trained DNN. In order to force the two filter sets similar, we propose a new training algorithm which is achieved by adding a ``cluster loss" on the original loss function. The network can learn compact representations during training with backpropagation algorithm. As  shown in Fig \ref{filter_cluster},our method consists of three main steps: (1) Given a DNN, we define the clusters which each filter belong to and train the network with our proposed training algorithm. (2) After training, filters in each cluster are similar and one of them can be removed, besides, the corresponding filter channels can also be pruned. (3) At last, we fine-tune the pruned network to compensate the performance degradation.

We evaluate our proposed method on two bechmark datasets (CIFAR10 and CIFAR100) and demonstrate its competitive performance compared with state-of-the-art approaches. For VGG-16, our method shows 2$\times$ speedup without loss in accuracy. With WRN-16-4, it achieves about 3.4$\times$  and 1.74$\times$ speedup within 1\% accuracy drop on CIFAR10 and CIFAR100, respectively.

\section{Related Work}

As one of the most popular methods for accelerating network inference, pruning has been widely studied recently. In the earlier work, Optimal Brain Damage \cite{lecun1990optimal} prunes unimportant weights to reduce the number of parameters and prevent over-fitting. Recently, Han \emph{et al.} \cite{han2015learning} prune the weights which are below a threshold results in a very sparse model with no loss of accuracy. But such a non-structured sparse model has limitations of accelerating inference without specific hardware \cite{luo2017thinet}. Latter, researchers focus on filter level pruning which can not only reduce the memory footprint dramatically but also speedup network inference by any off-the-shelf library. Li \emph{et al.} \cite{li2016pruning} prune the less useful filters based on sum of absolute weights directly.

But the filter of small magnitude does not mean it is not important. Thus, the methods based on the information of activations are studied. Hu \emph{et al.} \cite{hu2016network} calculate the sparsity of activations after the ReLU function and prune the corresponding filters if the sparsity is high. Molchanov \emph{et al.} \cite{molchanov2016pruning} adopt a first-order Taylor expansion to approximate the change to loss function induced by pruning each filter.   He \emph{et al.} \cite{he2017channel} prune channels by a LASSO regression based channel selection and least square reconstruction. Yu \emph{et al.} \cite{yu2017nisp} prune the entire  network jointly to minimize the reconstruction error of important responses in the ``final response layer" and propagate the importance scores of final responses to every neuron.

 Besides the above methods which prune filters on a pre-trained network,  several methods add regularization or other modifications  during training.  For example,  Liu \emph{et al.} \cite{liu2017learning}    enforce channel sparsity by imposing L1 regularization on the scaling factors in batch normalization. McDanel \emph{et al.} \cite{mcdanel2017incomplete} utilize incomplete dot products to dynamically adjust the number of input channels used in each layer. Zhou \emph{et al.} \cite{zhou1226} introduce a scaling factor to each filter to weaken the weights step by step and prune the filters after training.

Meanwhile, other methods are well explored to compress networks. Knowledge distillation \cite{hinton2015distilling} first trains a big network (\emph{i.e.,} teacher network) and then trains a shallow one (\emph{i.e.,} student network) to mimic the output distributions of the teacher. Network quantization reduces the number of bits for representing each parameter and some low-bit networks are proposed \cite{courbariaux2016binarized}\cite{li2016ternary}\cite{zhuxiaotian}. Low-rank factorization decompose weights into several pieces \cite{howard2017mobilenets}\cite{lebedev2014speeding}. Note that our pruning method can be integrated with the above techniques to achieve a more compact and efficient model.

\begin{figure}[t]
\centering
\includegraphics[scale = 0.9]{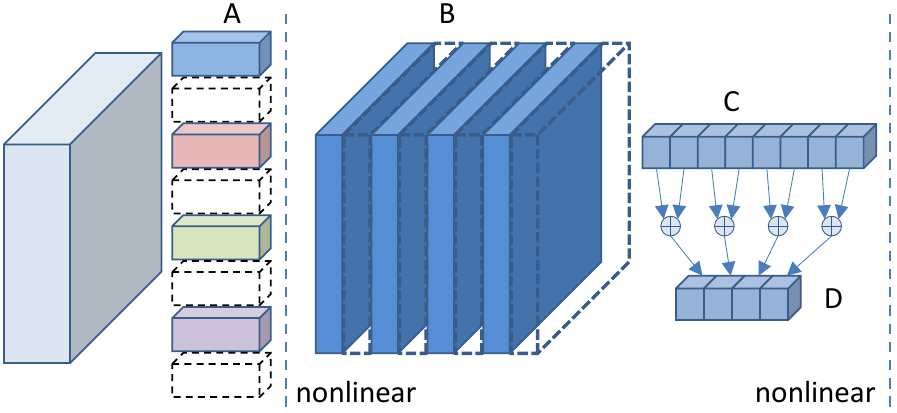}
\vspace{-0.3cm}
\caption{Pruning similar filters in a convolutional layer. Whenever two filters sets are similar, we remove one of them for A, while the corresponding channels of the output activations B can also be pruned.Moreover, the channels of filter in next layer can also be removed, but we need to add one channel to the other in each sets (\emph{i.e.,} from C to D) as shown in Eq (\ref{c}).  }
\label{filter_prune}
\end{figure}
\section{Our Method}
In this section, we first show how we prune the similar filters for a singe layer, then propose a new training algorithm for forcing filters in each cluster to be similar. Finally, we analyze the acceleration and compression of our method.

\subsection{Pruning Similar Filters}
We use a triplet $\langle Y_i, K_i, \ast \rangle$ to denote the convolution process in layer \emph{i}, where $Y_i \in \mathbb{R}^{H\times W\times M}$ is the input tensor, which has $M$ channels, $H$ rows and $W$ columns. And $K_i \in \mathbb{R}^{d\times d\times M\times N}$ is a filter bank with $d\times d$ kernel size, $\ast$ denote the 3D convolution operation which maps $Y_i$ to $Y_{i+1}$ using $K_i$, where $Y_{i+1}$ is the input tensor in layer $i+1$ (which is also the output tensor of layer $i$), Note that the fully connected operation is a special case of convolution operation.

Formally, the $j^{th}$ channel of $Y_{i+1}$ can be computed with the $j^{th}$ filter and the input tensor $Y_{i}$ using the 2D convolution operation $\otimes$ as follows :
\vspace{-0.5cm}
\begin{eqnarray}\label{a}
\textbf{$y^j_{i+1}$} &=& h(\textbf{$k_i^{j}$} \ast \textbf{$y_{i}$})= h(\sum_{m=1}^{M}{\textbf{$k_i^{jm}$} \otimes \textbf{$y^m_{i}$}})   \nonumber\\
&=& h(\textbf{$k_i^{j1}$} \otimes \textbf{$y^1_{i}$} +  \textbf{$k_i^{j2}$} \otimes \textbf{$y^2_{i}$} + \ldots +\textbf{$k_i^{jM}$} \otimes \textbf{$y^M_{i}$}),
\end{eqnarray}
where $h(\cdot)$ is the nonlinear function, such as sigmoid or ReLU, $y^m_{i}$ is the $m^{th}$ channel and $k_i^{jm}$  is the $m^{th}$ channel of the $j^{th}$ filter in layer $i$.   Note that we ignore the corresponding bias for simplicity. In the same way, $y^j_{i}$ can be computed as:
\vspace{-0.2cm}
\begin{eqnarray}\label{b}
\textbf{$y^j_{i}$} &=& h(\textbf{$k_{i-1}^{j}$} \ast \textbf{$y_{i-1}$}).
\end{eqnarray}

Now let us suppose that $k_{i-1}^1=k_{i-1}^2$. This means that the corresponding feature maps are the same, \emph{e.g.,}  $y^1_{i}=y^2_{i}$ according to Eq (\ref{b}). For $j^{th}$ filter, replacing $k_{i}^{j2}$ by $k_{i}^{j1}$ in Eq (\ref{a}), we get the Eq (\ref{c}).
\vspace{-0.2cm}
\begin{eqnarray}\label{c}
\textbf{$y^j_{i+1}$}=h(\textbf{$(k_i^{j1}+k_i^{j2})$} \otimes \textbf{$y^1_{i}$} +  \textbf{0} \otimes \textbf{$y^2_{i}$} + \ldots +\textbf{$k_i^{jM}$} \otimes \textbf{$y^M_{i}$}),
\end{eqnarray}
where $\textbf{0}$ is the channel of all zero-value.
This means that whenever two filter sets in layer $i-1$ are equal, we can prune one of them safely. At the same time,  we modify the filter channels in layer $i$ (\emph{i.e.,} for each filter, we add one channel to the other). Fig. \ref{filter_prune} exhibits the operation clearly.
\subsection{Proposed Training Algorithm}
However, two filter sets could never be exactly equal in a trained CNN, and it's hard to find such two similar filter sets. To cope with it, we propose a new training algorithm to force several two filter sets to be equal. Let $K$ denote the weights in the network and $f(K)$ is our loss function, the optimization target can be formulated as:
\vspace{-0.5cm}
\begin{eqnarray}\label{d}
f(K) &=& E(K)+\delta R(K)+\lambda \sum_{i=1}^{L}{\sum_{t=1}^{T}{  \Arrowvert k_i^t-c_i^t \Arrowvert ^2    }}, \\
c_i^t &=& \frac{1}{\arrowvert S_i^t \arrowvert}\sum_{k_i^t\in S_i^t}{k_i^t},
\end{eqnarray}
where $E(K)$ and $R(K)$ refer to the cross-entropy loss and the regularization loss, respectively, $\delta$ and $\lambda$ are the tunable parameters to balance the loss terms, $\Arrowvert k_i^t-c_i^t \Arrowvert ^2$ denotes the ``cluster loss" of the $t^{th}$ cluster in layer $i$. Thus, the filters in each cluster $S_i^t$ are forced to be similar using the cluster loss and just one filter is kept after training.

If the clusters are changing during training, it may cause the training process difficult due to the high dimension of each filter and the non-balanced clusters.
Therefore, in this work, we can define the clusters $S_i^t$  which the filters belong to firstly and keep the clusters not change online. And the number of cluster in each layer is specified with the compression rate. 

Specifically, let $N$ denote the number of filters in layer $i$ and $N\times p$ is the number of filters are pruned ($\frac{1}{1-p}$ is the compression rate). We restrict $p\leq 0.5$,  because the performance degrades severely when over half of filters are pruned. The size of each cluster $\arrowvert S_i^t \arrowvert$  is no larger than two for balance.  Thus, there are $N\times (1-p)$ clusters, where each $N\times p$ cluster contains two filters and each $N\times (1-2p)$ cluster contains just one filter. Note that one of  the filters in each cluster whose size is larger than one is pruned and the clusters of size equal to one are preserved.

\textbf{Analysis of Compression and Acceleration.}
According to Eq (\ref{c}) and Fig. \ref{filter_prune},  if a filter in layer $i$ is removed, its corresponding channel of filters in layer $i+1$ is also discarded. The size of each feature map is kept the same.
Since we preserve $N_i\times (1-p_i)$ filters in layer $i$. The speedup ratio $r_{s_i}$ and compression ratio $r_{c_i}$ of the pruned network for layer $i$ compared to the original network can be computed by $ r_{s_i} =r_{c_i} = (1-p_{i-1})(1-p_i)$.
Note that the feature maps in layer $i$ are also compressed $1-p_i$ times which saves a large proportion of runtime memory.

\begin{figure}[htbp]
\begin{minipage}[b]{0.48\linewidth}
  \centering
\includegraphics[scale = 0.32]{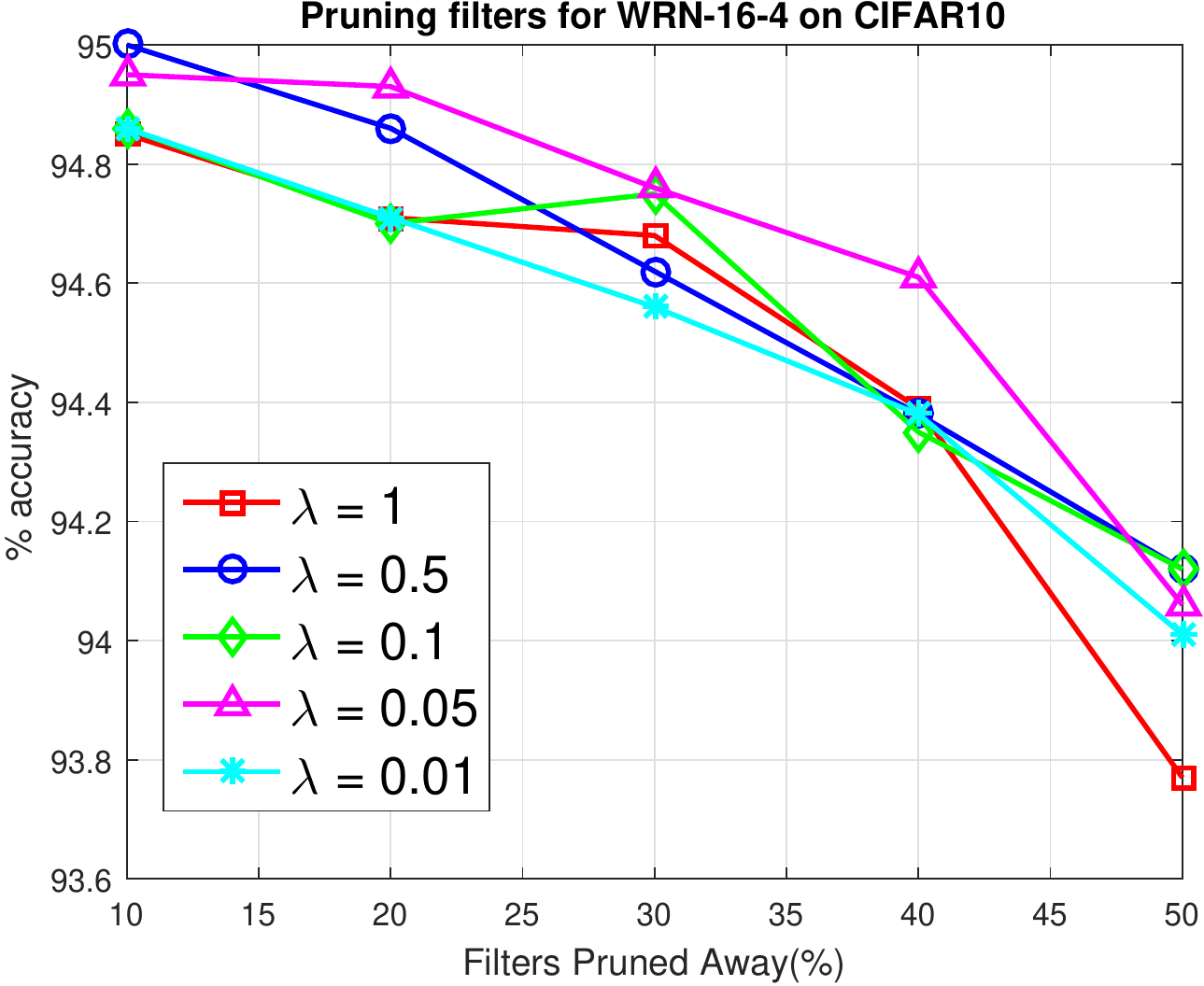}
\end{minipage}
\begin{minipage}[b]{.48\linewidth}
  \centering
\includegraphics[scale = 0.32]{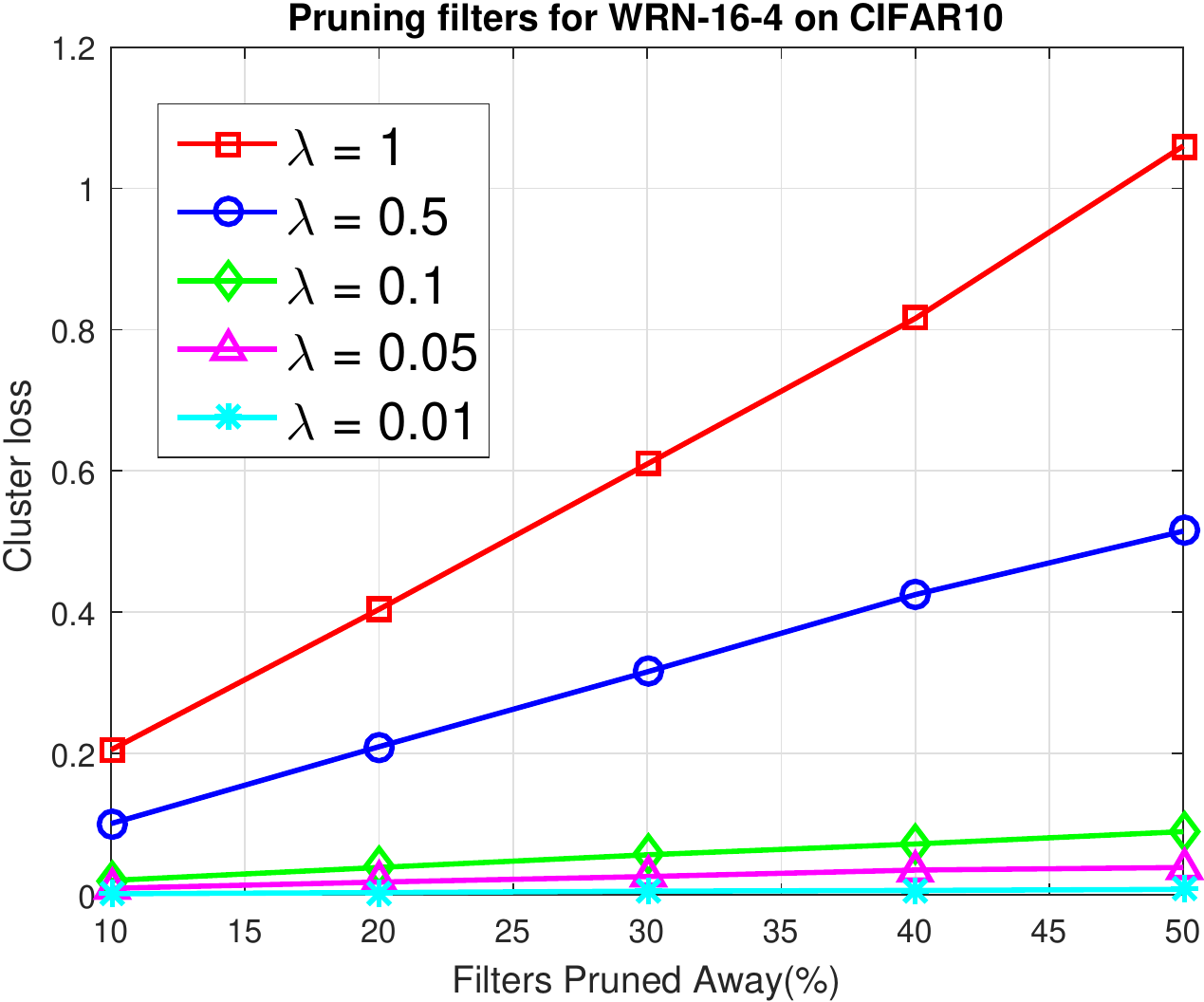}
\end{minipage}
\caption{Accuracy after pruning filters (left) and the final ``cluster loss"  under different pruned ratio and different $\lambda$ (Eq (\ref{d})) for the WRN-16-4 model on CIFAR10 (right). }
\label{lambda}
\end{figure}
\section{Experiments}
We evaluate our proposed method on several typical networks and two datasets (CIFAR-10 and CIFAR-100) \cite{alex2009learning}. We use TensorFlow \cite{abadi2016tensorflow} framework to implement network pruning and evaluate on Nvidia GTX 1080Ti GPU.

\begin{figure*}[!htb]
\begin{minipage}[b]{0.247\linewidth}
  \centering
\includegraphics[scale = 0.343]{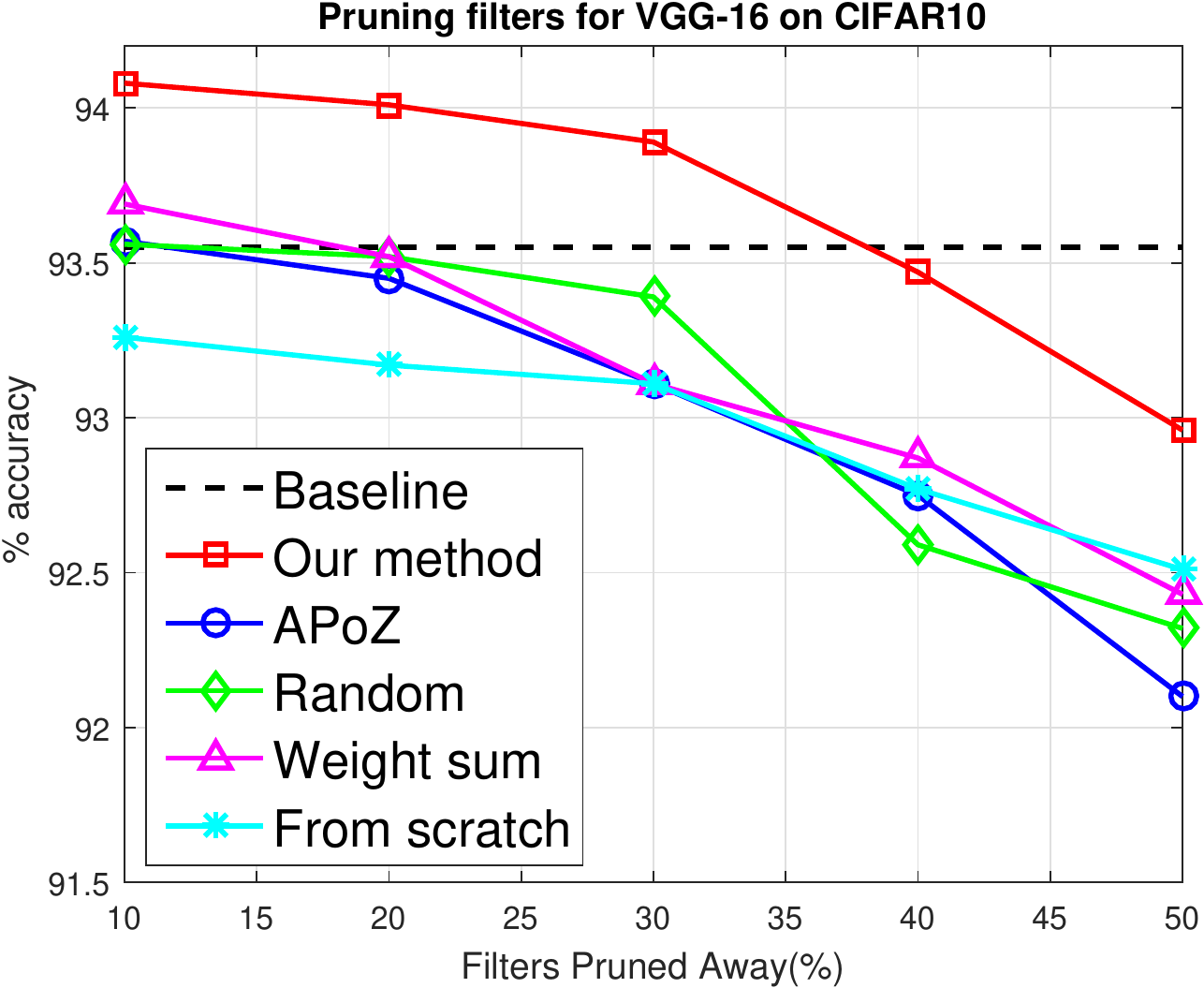}
\end{minipage}
\begin{minipage}[b]{.247\linewidth}
  \centering
\includegraphics[scale = 0.343]{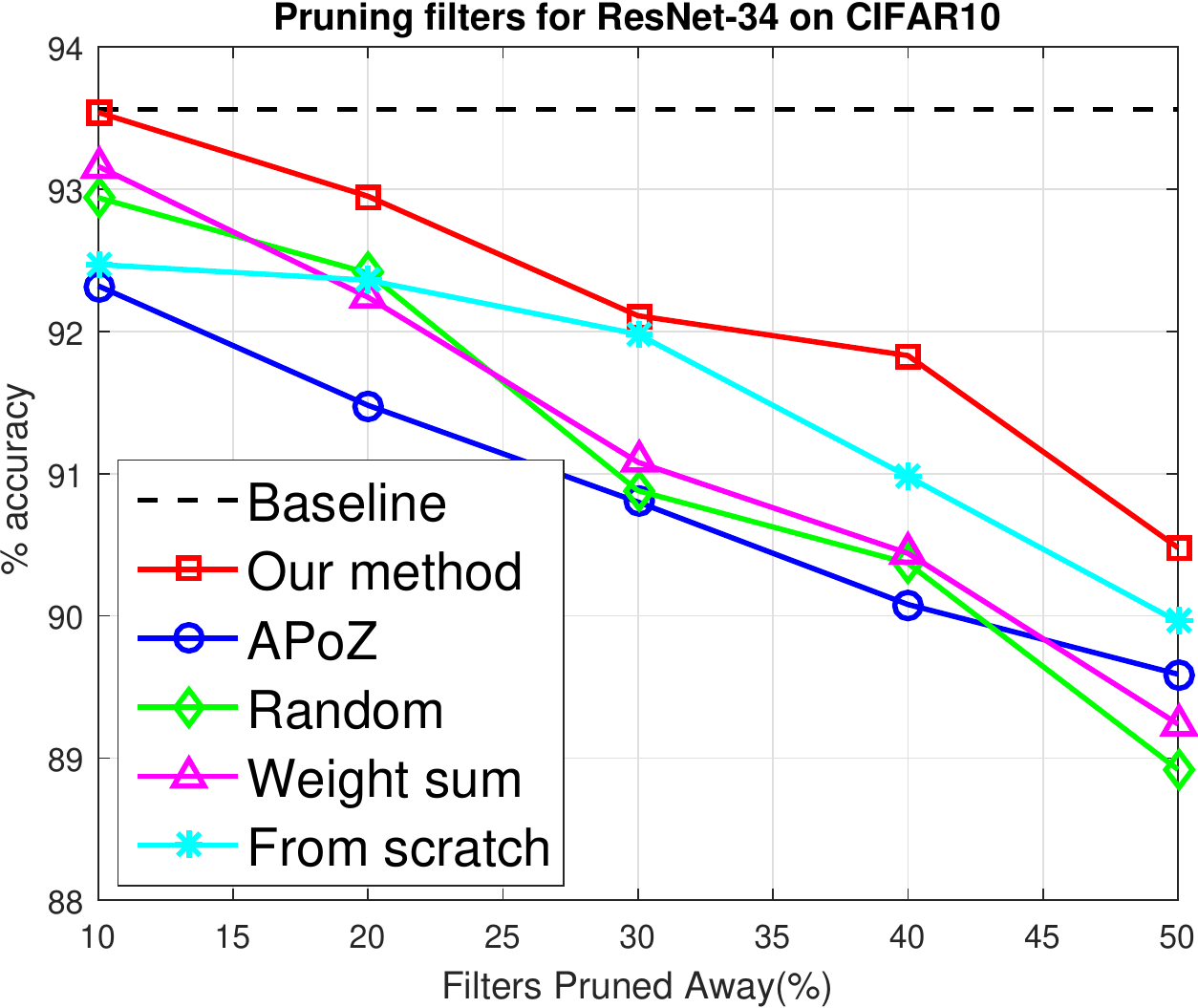}
\end{minipage}
\begin{minipage}[b]{0.247\linewidth}
  \centering
\includegraphics[scale = 0.343]{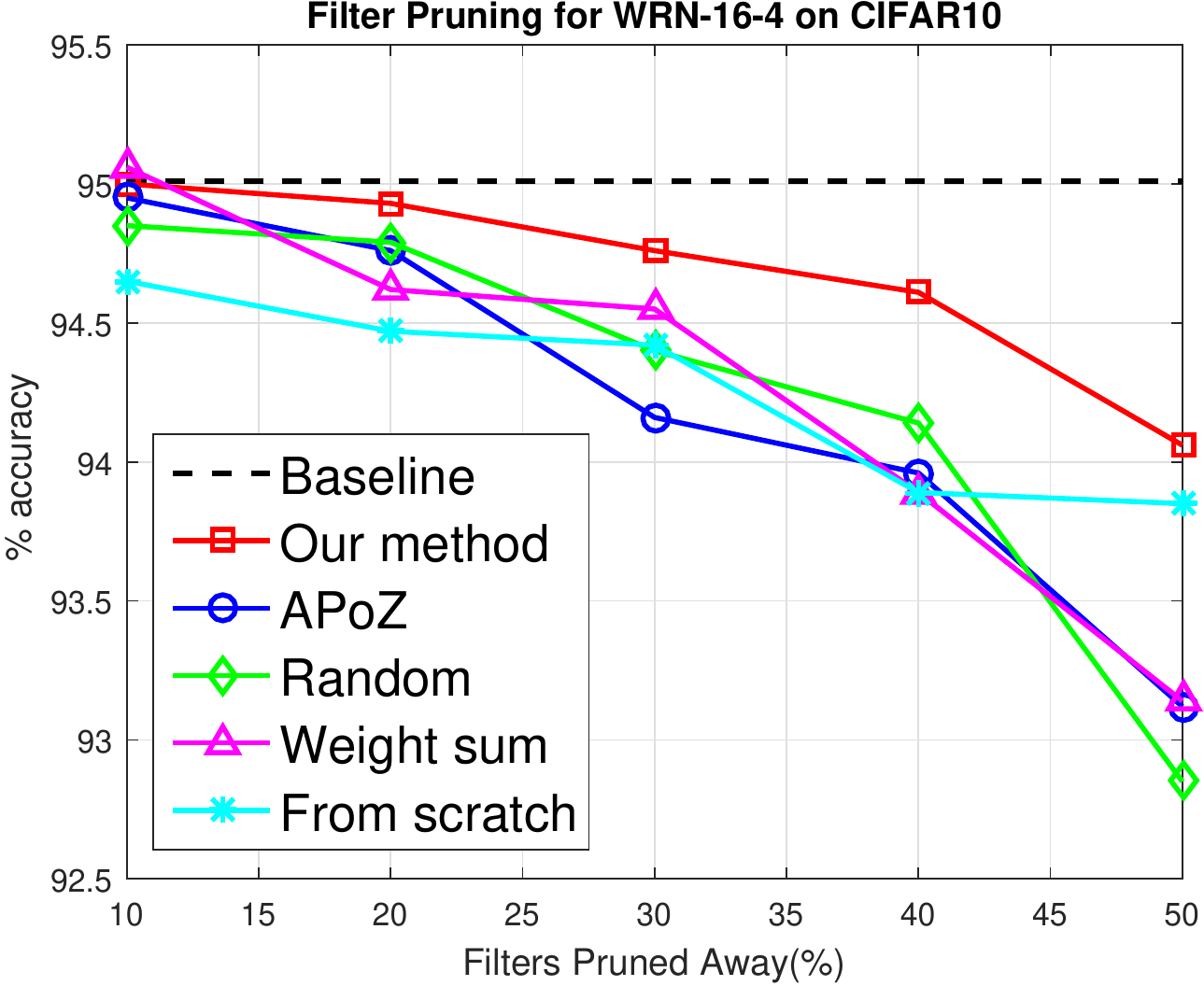}
\end{minipage}
\begin{minipage}[b]{0.247\linewidth}
  \centering
\includegraphics[scale = 0.343]{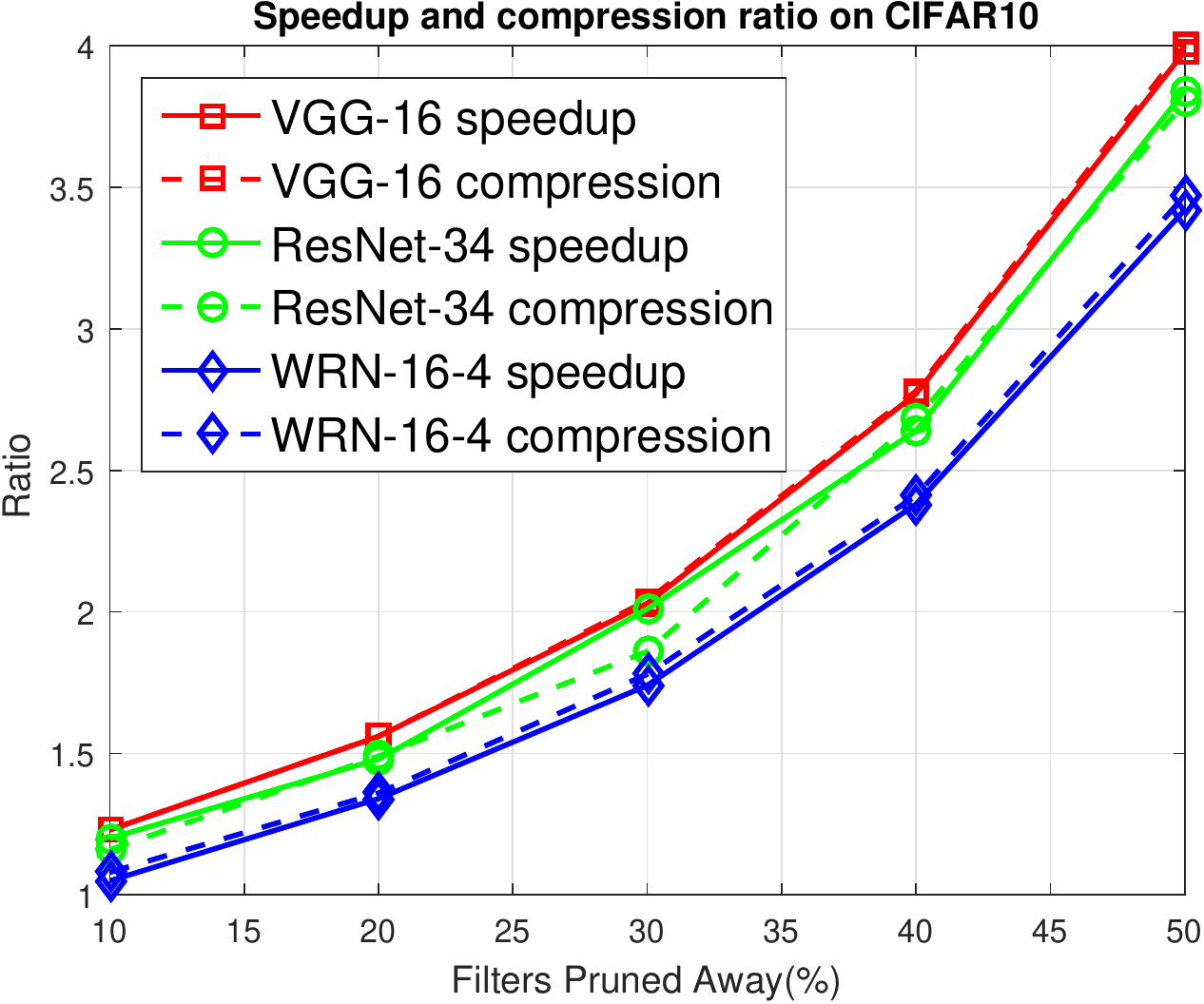}
\end{minipage}

\vspace{0.2cm}

\begin{minipage}[b]{0.247\linewidth}
  \centering
\includegraphics[scale = 0.343]{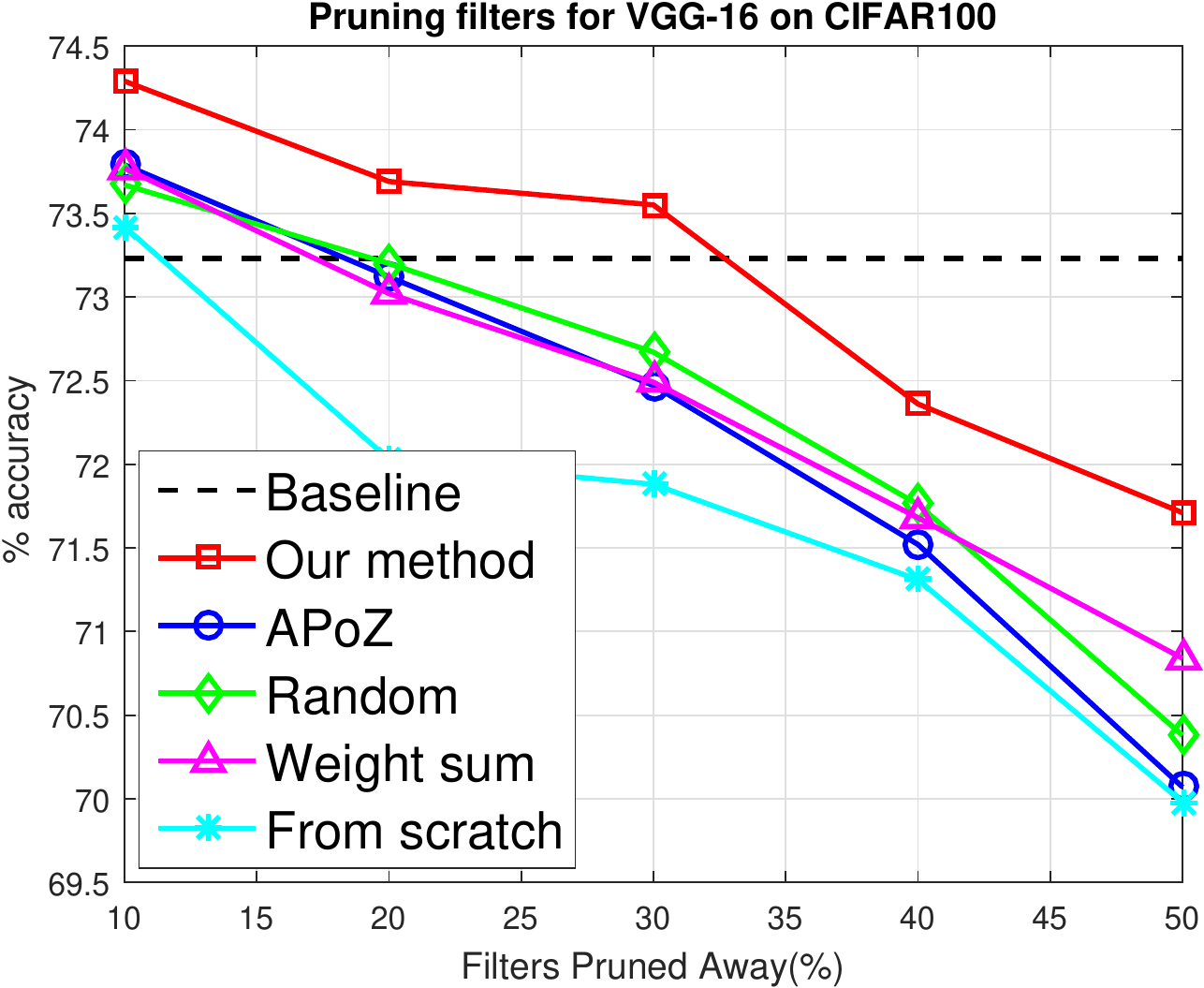}
\end{minipage}
\begin{minipage}[b]{.247\linewidth}
  \centering
\includegraphics[scale = 0.343]{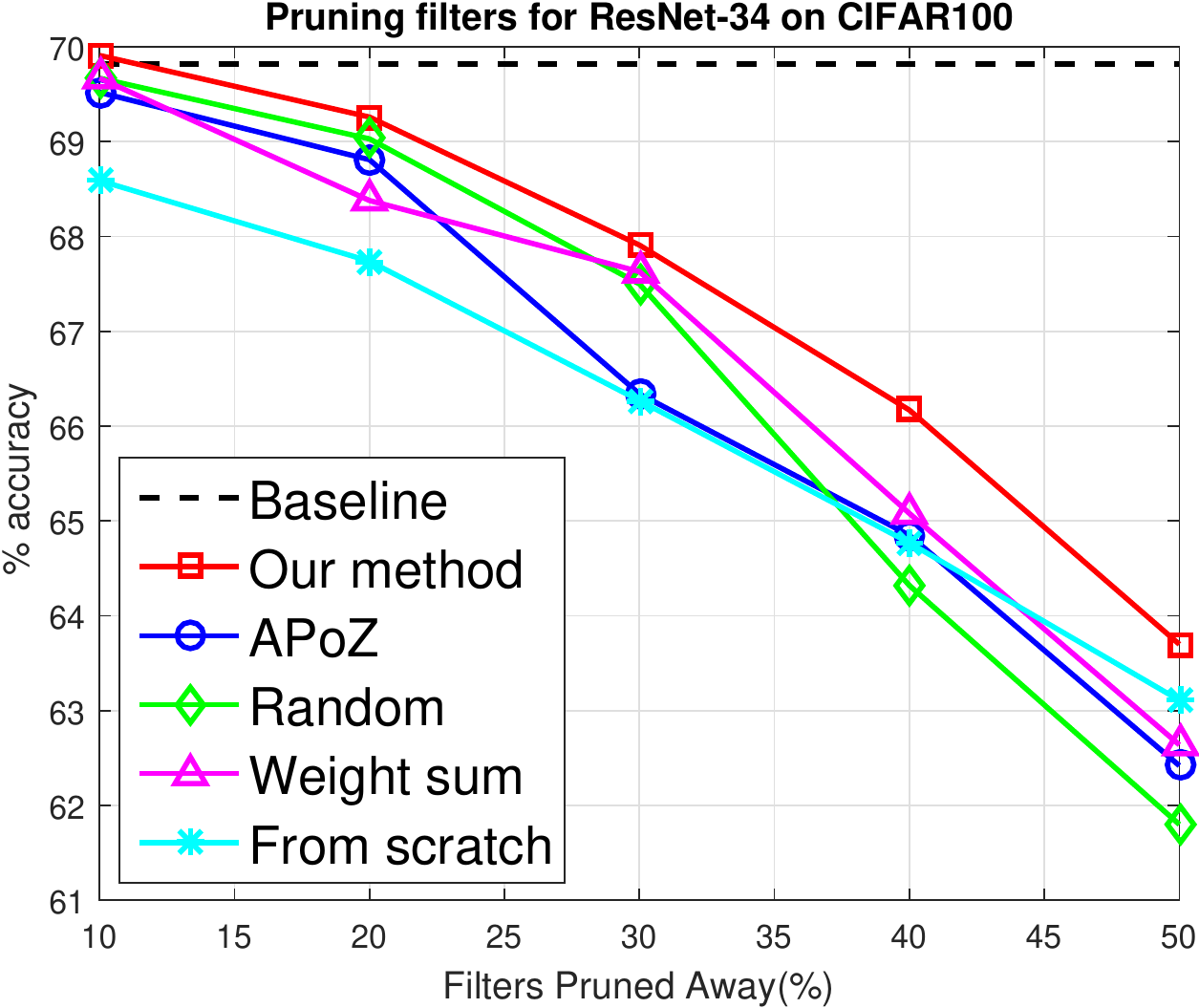}
\end{minipage}
\begin{minipage}[b]{0.247\linewidth}
  \centering
\includegraphics[scale = 0.343]{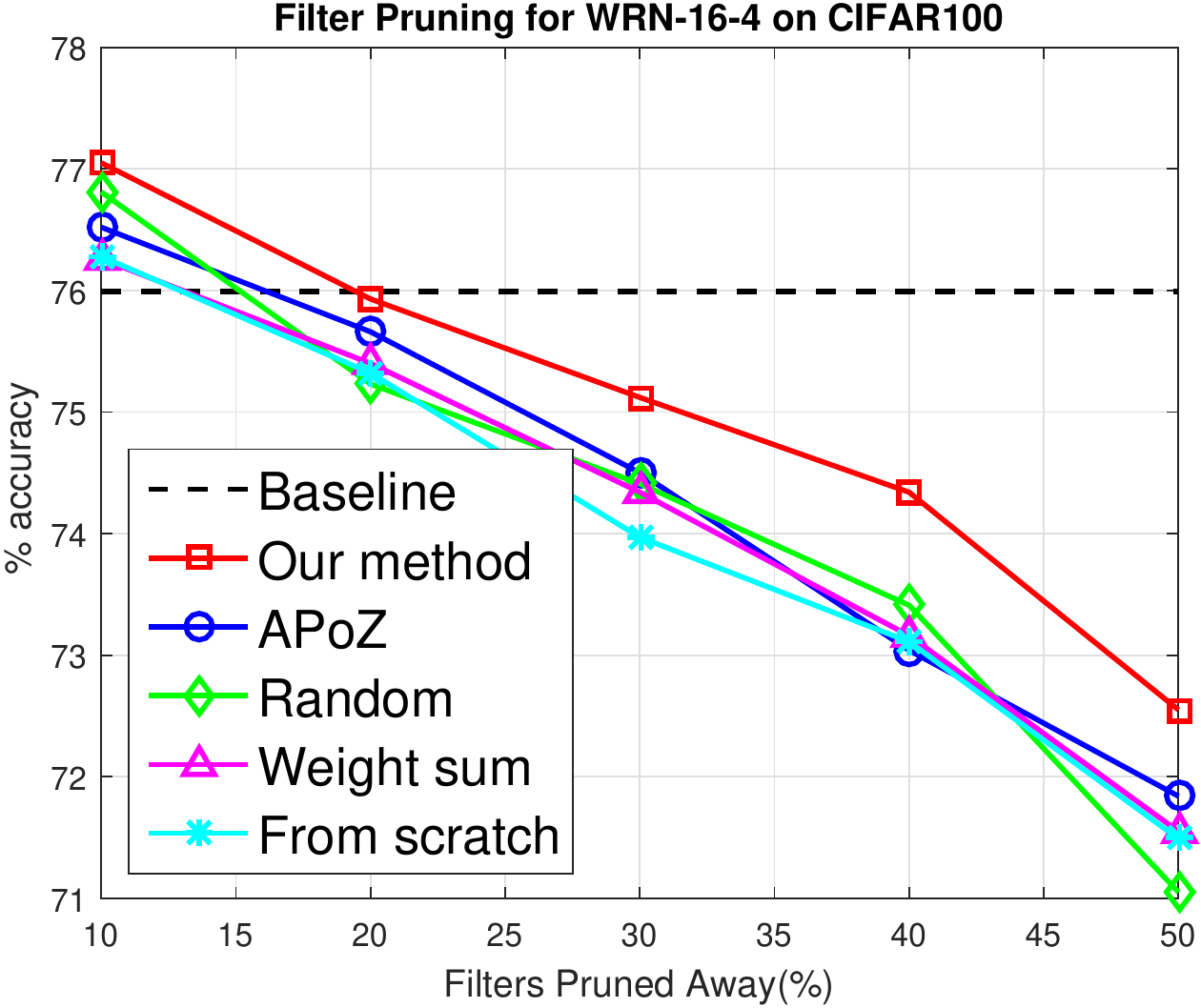}
\end{minipage}
\begin{minipage}[b]{0.247\linewidth}
  \centering
\includegraphics[scale = 0.343]{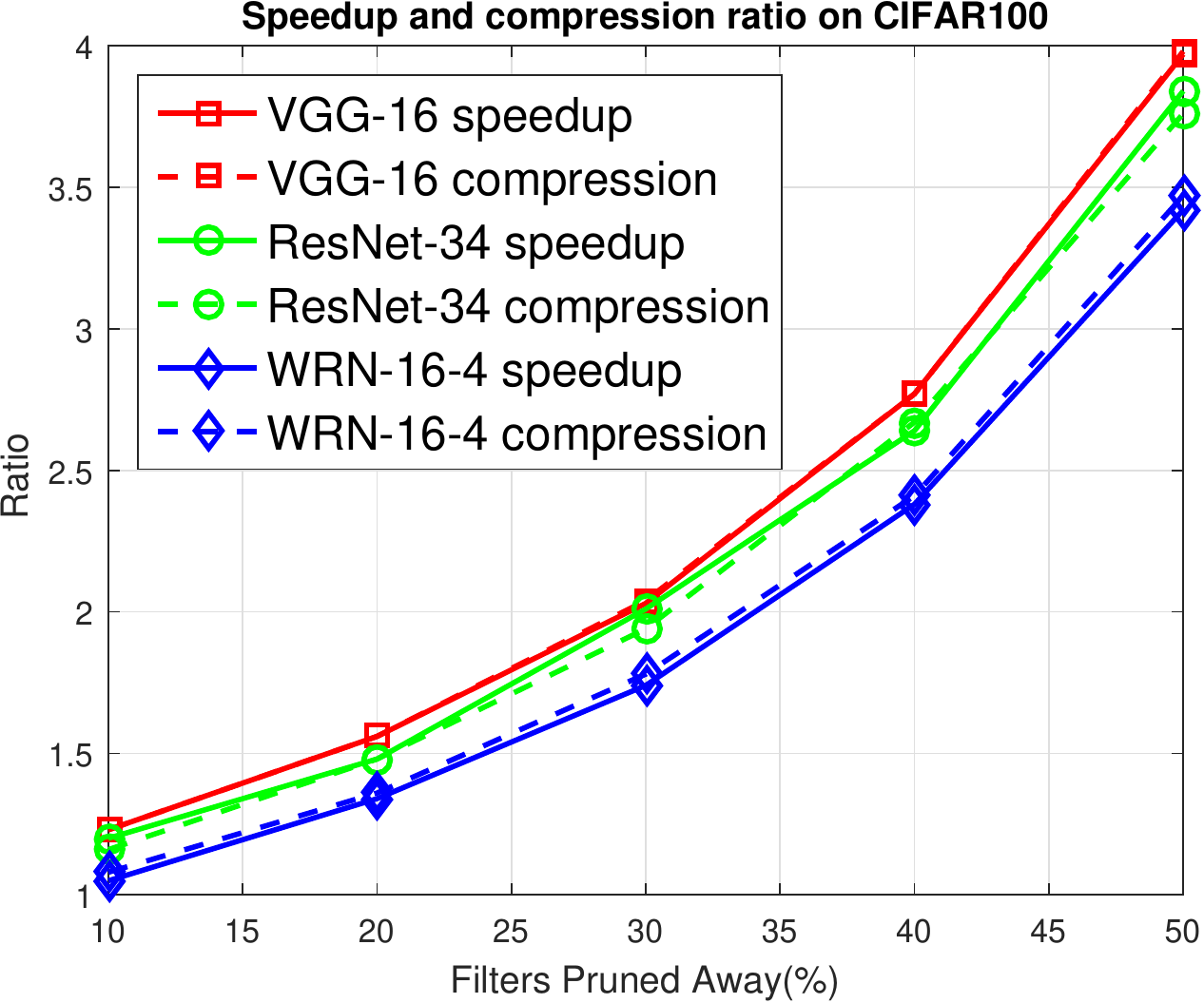}
\end{minipage}

\vspace{-0.3cm}
\caption{A comparison of the performance  with different filter selection criteria and different pruned ratio. The first row is the VGG-16, ResNet-34 and WRN-16-4 network on CIFAR10, respectively. The second row is the models on CIFAR100. Our approach is consistently better (larger is better). Besides, the last column is the speedup and compression ratio of models.}
\label{acc}
\end{figure*}
\subsection{Implementation Details and  filter selection criteria}
CIFAR10 and CIFAR100 both consist of a training set of 50000 and a test set of 10000 color images of size $32 \times 32$ with 10 classes and 100 classes, respectively. The training images are padded by 4 pixels and randomly flipped.

We evaluate three DNNs (\emph{i.e.,} VGG-16, ResNet-34 and WRN-16-4) on the two datasets, respectively. All networks are trained using SGD with batch size 128 and 300 epochs. The weight decay is 0.0005 and momentum is 0.9. The other hyper-parameters for the models are in  the following.
(1)The VGG-16 network is derived from \cite{li2016pruning}. The initial learning rate is set to 0.02, and is divided by 5 at each 60 epochs. We get the baseline accuracy on CIFAR10 and CIFAR100 of 93.55\% and 73.23\%, respectively. (2) The ResNet-34 model replaces shortcut layer with a $1 \times 1$ convolutional layer of  ResNet-32 \cite{dong2017more}. The initial learning rate is set to 0.1, and is divided by 5 at each 60 epochs. We get the baseline accuracy on CIFAR10 and CIFAR100 of 93.56\% and 69.82\%, respectively. (3) The WRN-16-4 network is adopted from \cite{zagoruyko2016wide}. The initial learning rate is set to 0.2, and is divided by 5 at each 60 epochs. We get the baseline accuracy on CIFAR10 and CIFAR100 of 95.01\% and 75.99\%, respectively.

In addition, we compare our method with other state-of-the-art criteria, including (1) \textbf{Weight sum} \cite{li2016pruning}. This criterion measure the importance of a filter in each layer by calculating the sum of its absolute weights ($i.e.,$ $\sum{\arrowvert k_i^j\arrowvert}$); (2) \textbf{Average Percentage of Zeros (APoZ)} \cite{hu2016network}. APoZ measures the percentage of zero activations of a filter after the ReLU mapping. The APoZ of the $c^{th}$ neuron is $\frac{\sum\nolimits_n^{\hat{N}}{\sum{I(L_c(n)==0)}}}{\hat{N}\times N}$, where $\hat{N}$ is the number of training examples (we set $\hat{N}=50000$); (3) \textbf{Randomly pruning}. During pruning, it randomly selects and removes filters. Moreover, we also compare the methods with \textbf{Train from scratch} which trains a model from scratch, where the model has the same structure as the pruned network.

\subsection{The effect of $\lambda$}
%
Firstly, we explore the influence of $\lambda$ (Eq (\ref{d})) on the performance of our pruned method. We set the pruned ratio the same in every layer of the WRN-16-4 network. Fig. \ref{lambda} demonstrates the accuracy of the pruned model with different pruned ratio and  $\lambda$ and the ``cluster loss" after training the model with our proposed training algorithm.
As expected, cluster loss increases as filter pruned ratio and $\lambda$ increases.
In addition, the losses are always small, so one filter in each cluster can be pruned with minimal  degradation of accuracy. The performance of pruned network is similar with different $\lambda$. In the rest of experiment, we set the $\lambda$ to 0.05.

\subsection{Comparison with other filter selection criteria}
Fig. \ref{acc} shows the pruned results for VGG-16, ResNet-34 and WRN-16-4 on CIFAR10 and CIFAR100 with different filter selection  criteria and different pruned ratios. Our approach is consistently better than other filter selection criteria under different pruned ratio. The method based on data (\emph{i.e.,} APoZ) is similar to other data-independent approaches. This may because the number of filters pruned is small and the gap of these methods is not obvious.  Training the pruned network from scratch is not always worse than other methods especially when the model is deep and wide. We can accelerate VGG-16 network 2$\times$ without loss in accuracy and speedup WRN-16-4 model on CIFAR10 about 3.4$\times$ with 1\% degradation on accuracy.  But the ResNet-34 is hard to compress, which may because the model is already compact and efficient.

\begin{center}
\begin{table}[t]
\small
\caption{A comparison of speedup  with other compression methods. Values in parentheses are the increased error.} \label{cp}
\begin{tabular}{|c|c|c|c|}
  \hline
  Dataset & Model &  Error (\%) & Speedup  \\
  \hline
\multicolumn{1}{|c|}{\multirow{5}{*}{CIFAR10}} &  VGG-16 \cite{li2016pruning}   &   6.75 (-0.15)    &  1.52$\times$    \\
& VGG-16 \cite{liu2017learning}   &   6.34 (-0.14)    &  2.04$\times$   \\
& \textbf{VGG-16 (ours)}                                  &   \textbf{6.45 (-0.21)}    &  \textbf{2.77$\times$}  \\
\cline{2-4}
&  ResNet-56 \cite{li2016pruning} &  6.96 (-0.02)  & 1.38$\times$     \\
& \textbf{ResNet-58 (ours)}                              &  \textbf{6.18 (-0.01)}          &  \textbf{1.50$\times$}    \\
  \hline
\multicolumn{1}{|c|}{\multirow{4}{*}{CIFAR100}} &  VGG-16 \cite{liu2017learning}   &   26.74 (-0.22)   &  1.59$\times$           \\
&  \textbf{VGG-16 (ours)}                                &    \textbf{26.77 (-0.32)}  &  \textbf{2.03$\times$}       \\
\cline{2-4}
&  WRN-16-4 \cite{dong2017more} &  24.53 (+0.30)  & 1.18$\times$     \\
&   \textbf{WRN-16-4 (ours)}                                & \textbf{24.01 (+0.06)}  & \textbf{1.34$\times$}    \\
  \hline
\end{tabular}
\end{table}
\end{center}
\vspace{-1.5cm}
\subsection{Comparison with other compression methods}
Besides filter pruning methods, we compare the acceleration of our approach with other network compression methods in Table \ref{cp}. In general, different layer has different importance and  sparsity \cite{dong2017more}, and the method based training \cite{liu2017learning} can automatically find it. Even though, our approach can outperform other methods.

\section{Conclusion}
In this work, we introduce the cluster loss on the original loss function to force filters in each cluster to be similar during training, and prune one filter in every cluster safely. The compact model is inference efficient and requires no special hardware.
Extensive experiments on two datasets demonstrate the competitive performance of our proposed  method.
 In the future, we would like to evaluate our method on larger dataset and more vision tasks.

\bibliographystyle{IEEEbib}
\bibliography{icip18}

\begin{thebibliography}{10}

\bibitem{liu20163d}
Xiabing Liu, Wei Liang, Yumeng Wang, Shuyang Li, and Mingtao Pei,
\newblock ``3d head pose estimation with convolutional neural network trained
  on synthetic images,''
\newblock in {\em ICIP}, 2016.

\bibitem{khalaf2016convolutional}
Aya~F Khalaf, Inas~A Yassine, and Ahmed~S Fahmy,
\newblock ``Convolutional neural networks for deep feature learning in retinal
  vessel segmentation,''
\newblock in {\em ICIP}, 2016.

\bibitem{sun2016scalable}
Shaoyan Sun, Wengang Zhou, Qi~Tian, and Houqiang Li,
\newblock ``Scalable object retrieval with compact image representation from
  generic object regions,''
\newblock {\em ACM TOMM}, vol. 12, no. 2, pp. 29, 2016.

\bibitem{lecun1990optimal}
Yann LeCun, John~S Denker, and Sara~A Solla,
\newblock ``Optimal brain damage,''
\newblock in {\em NIPS}, 1990.

\bibitem{han2015learning}
Song Han, Jeff Pool, John Tran, and William Dally,
\newblock ``Learning both weights and connections for efficient neural
  network,''
\newblock in {\em NIPS}, 2015.

\bibitem{zhou1226}
Zhengguang Zhou, Wengang Zhou, Richang Hong, and Houqiang Li,
\newblock ``Online filter weakening and pruning for efficient convnets,''
\newblock in {\em ICME}, 2018.

\bibitem{he2017channel}
Yihui He, Xiangyu Zhang, and Jian Sun,
\newblock ``Channel pruning for accelerating very deep neural networks,''
\newblock in {\em ICCV}, 2017.

\bibitem{hu2016network}
Hengyuan Hu, Rui Peng, Yu-Wing Tai, and Chi-Keung Tang,
\newblock ``Network trimming: A data-driven neuron pruning approach towards
  efficient deep architectures,''
\newblock {\em arXiv preprint arXiv:1607.03250}, 2016.

\bibitem{li2016pruning}
Hao Li, Asim Kadav, Igor Durdanovic, Hanan Samet, and Hans~Peter Graf,
\newblock ``Pruning filters for efficient convnets,''
\newblock {\em arXiv preprint arXiv:1608.08710}, 2016.

\bibitem{liu2017learning}
Zhuang Liu, Jianguo Li, Zhiqiang Shen, Gao Huang, Shoumeng Yan, and Changshui
  Zhang,
\newblock ``Learning efficient convolutional networks through network
  slimming,''
\newblock {\em arxiv preprint}, vol. 1708, 2017.

\bibitem{luo2017thinet}
Jian-Hao Luo, Jianxin Wu, and Weiyao Lin,
\newblock ``Thinet: A filter level pruning method for deep neural network
  compression,''
\newblock {\em arXiv preprint arXiv:1707.06342}, 2017.

\bibitem{molchanov2016pruning}
Pavlo Molchanov, Stephen Tyree, Tero Karras, Timo Aila, and Jan Kautz,
\newblock ``Pruning convolutional neural networks for resource efficient
  transfer learning,''
\newblock {\em arXiv preprint arXiv:1611.06440}, 2016.

\bibitem{srinivas2015data}
Suraj Srinivas and R~Venkatesh Babu,
\newblock ``Data-free parameter pruning for deep neural networks,''
\newblock {\em arXiv preprint arXiv:1507.06149}, 2015.

\bibitem{yu2017nisp}
Ruichi Yu, Ang Li, Chun-Fu Chen, Jui-Hsin Lai, Vlad~I Morariu, Xintong Han,
  Mingfei Gao, Ching-Yung Lin, and Larry~S Davis,
\newblock ``Nisp: Pruning networks using neuron importance score propagation,''
\newblock {\em arXiv preprint arXiv:1711.05908}, 2017.

\bibitem{mcdanel2017incomplete}
Bradley McDanel, Surat Teerapittayanon, and HT~Kung,
\newblock ``Incomplete dot products for dynamic computation scaling in neural
  network inference,''
\newblock {\em arXiv preprint arXiv:1710.07830}, 2017.

\bibitem{hinton2015distilling}
Geoffrey Hinton, Oriol Vinyals, and Jeff Dean,
\newblock ``Distilling the knowledge in a neural network,''
\newblock {\em arXiv preprint arXiv:1503.02531}, 2015.

\bibitem{courbariaux2016binarized}
Matthieu Courbariaux, Itay Hubara, Daniel Soudry, Ran El-Yaniv, and Yoshua
  Bengio,
\newblock ``Binarized neural networks: Training deep neural networks with
  weights and activations constrained to+ 1 or-1,''
\newblock {\em arXiv preprint arXiv:1602.02830}, 2016.

\bibitem{li2016ternary}
Fengfu Li, Bo~Zhang, and Bin Liu,
\newblock ``Ternary weight networks,''
\newblock {\em arXiv preprint arXiv:1605.04711}, 2016.

\bibitem{zhuxiaotian}
Xiaotian Zhu, Wengang Zhou, and Houqiang Li,
\newblock ``Adaptive layerwise quantization for deep neural network
  compression,''
\newblock in {\em ICME}, 2018.

\bibitem{howard2017mobilenets}
Andrew~G Howard, Menglong Zhu, Bo~Chen, Dmitry Kalenichenko, Weijun Wang,
  Tobias Weyand, Marco Andreetto, and Hartwig Adam,
\newblock ``Mobilenets: Efficient convolutional neural networks for mobile
  vision applications,''
\newblock {\em arXiv preprint arXiv:1704.04861}, 2017.

\bibitem{lebedev2014speeding}
Vadim Lebedev, Yaroslav Ganin, Maksim Rakhuba, Ivan Oseledets, and Victor
  Lempitsky,
\newblock ``Speeding-up convolutional neural networks using fine-tuned
  cp-decomposition,''
\newblock {\em arXiv preprint arXiv:1412.6553}, 2014.

\bibitem{alex2009learning}
Alex Krizhevsky and Geoffrey Hinton,
\newblock ``Learning multiple layers of features from tiny images,''
\newblock Tech. {R}ep., University of Toronto, 2009.

\bibitem{abadi2016tensorflow}
Mart{\'\i}n Abadi, Ashish Agarwal, Paul Barham, Eugene Brevdo, Zhifeng Chen,
  Craig Citro, Greg~S Corrado, Andy Davis, Jeffrey Dean, Matthieu Devin,
  et~al.,
\newblock ``Tensorflow: Large-scale machine learning on heterogeneous
  distributed systems,''
\newblock {\em arXiv preprint arXiv:1603.04467}, 2016.

\bibitem{dong2017more}
Xuanyi Dong, Junshi Huang, Yi~Yang, and Shuicheng Yan,
\newblock ``More is less: A more complicated network with less inference
  complexity,''
\newblock {\em arXiv preprint arXiv:1703.08651}, 2017.

\bibitem{zagoruyko2016wide}
Sergey Zagoruyko and Nikos Komodakis,
\newblock ``Wide residual networks,''
\newblock {\em arXiv preprint arXiv:1605.07146}, 2016.

\end{thebibliography}

\end{document}